# Evolution of Symbiosis in the Game of Life:

# Three Characteristics of Successful Symbiotes

Peter D. Turney[*]

## Abstract

In past work, we developed a computational model of the evolution of symbiotic entities (Model-S), based on Conway's Game of Life. In this article, we examine three trends that biologists have observed in the evolution of symbiotes. (1) *Management:* If one partner is able to control the symbiotic relation, this control can reduce conflict; thus, evolutionary selection favours symbiotes that have a manager. (2) *Mutualism:* Although partners in a symbiote often have conflicting needs, evolutionary selection favours symbiotes in which partners are better off together inside the symbiote than they would be as individuals outside of the symbiote. (3) *Interaction:* Repeated interaction among partners in symbiosis tends to promote increasing fitness due to evolutionary selection. We have added new components to Model-S that allow us to observe these three trends in runs of Model-S. The new components are analogous to the practice of staining cells in biology research, to reveal patterns that are not usually visible. When we measure the fitness of a symbiote by making it compete with other symbiotes, we find that fitter symbiotes have significantly more management, mutualism, and interaction than less fit symbiotes. These results confirm the trends observed in nature by biologists. Model-S allows biologists to study these evolutionary trends and other characteristics of symbiosis in ways that are not tractable with living organisms.

**Keywords:** Symbiosis, evolution, cellular automata, mutualism, interaction, management.

---

[*] Ronin Institute, 127 Haddon Place, Montclair, NJ 07043-2314, USA,

peter.turney@ronininstitute.org, 819-661-4625.

# 1 Introduction

Martin Gardner (1970) presented John Conway's Game of Life as three desiderata and three rules. The desiderata are as follows (Gardner, 1970, p. 120):

1. There should be no initial pattern for which there is a simple proof that the population can grow without limit.

2. There should be initial patterns that *apparently* do grow without limit.

3. There should be simple initial patterns that grow and change for a considerable period of time before coming to end in three possible ways: fading away completely (from overcrowding or from becoming too sparse), settling into a stable configuration that remains unchanged thereafter, or entering an oscillating phase in which they repeat an endless cycle of two or more periods.

The three rules are as follows (Gardner, 1970, p. 120):

1. Survivals. Every counter with two or three neighboring counters survives for the next generation.

2. Deaths. Each counter with four or more neighbors dies (is removed) from overpopulation. Every counter with one neighbor or none dies from isolation.

3. Births. Each empty cell adjacent to exactly three neighbors—no more, no fewer—is a birth cell. A counter is placed on it at the next move.

The basic premise of the current article is that we should take these desiderata and rules seriously, as a model for understanding biological life. If these rules capture something essential about life, then perhaps we can use them as a foundation for making an abstract model of biological organisms. In this article, we attempt to use the Game of Life as the basis for a model of symbiosis.

The word *symbiosis* comes from the Greek *symbíōsis*, meaning *living together*. Often symbiosis is understood as a mutually beneficial relationship, but biologists define symbiosis more generally as any long-term interaction between two or more different species. Following Douglas (2010), we call the members of a symbiote *partners*.

Martin and Schwab (2013) propose a set of terms based on whether two partners in a symbiotic relationship experience a beneficial effect (+), a harmful effect (-), or a neutral effect (0). This yields six types of symbiosis: neutralism (0/0), antagonism (-/-), amensalism (0/-), agonism (+/-), commensalism (+/0), and mutualism (+/+). It is often assumed that symbiotes are composed of two partners, but many symbiotes are composed of more partners. For example, the human body is estimated to contain 500 to 1000 species of symbiotic bacteria (Gilbert et al., 2018).

In previous publications (Turney, 2020; 2021a), we presented Model-S, a computational simulation of the evolution of symbiosis, based on a variation of John Conway's Game of Life (Gardner, 1970). A run of Model-S begins with a randomly generated population of single organisms (no symbiotes). An evolutionary algorithm allows the population to change over time, with mutation, asexual reproduction, sexual reproduction, selection, and symbiosis. As Model-S runs, the number of symbiotes in an organism increases. There is no fixed limit on the number of symbiotes in an organism.

There are several analogical relations between characteristics of biological life and characteristics of Model-S (Turney 2021a). In an abstract and simplified way, Model-S represents eight of the main concepts of biology, as listed in Table 1.

Table 1. The table shows eight correspondences between characteristics of biological life and Model-S. This table first appeared in an earlier article (Turney, 2021a), where more detailed explanations can be found.

| Biological Life | Model-S | Shared Characteristics |
| --- | --- | --- |
| genome, DNA, chromosomes | initial seed pattern, initial soup | static information, code |
| phenome, organism, development | playing the game, running the cellular automaton | dynamic development, growth |
| natural selection | tournament selection | selection |
| competition | the Immigration Game | fitness measure |
| reproduction with mutation, variation, sex | Layers 1, 2, and 3 of Model-S | reproduction with heritable variation |
| mutualism, cooperation | Layer 4, fusion of seed patterns | symbiosis |
| cells | still lifes, oscillators, spaceships, ashes | autopoiesis |
| multicellular organisms | multiple ashes grown from the same initial seed pattern | multicellularity |

Model-S was developed to provide insight into the major cooperative transitions in biological and cultural evolution (Maynard Smith and Szathmáry, 1995; Nolan and Lenski, 2010). In the current article, we show that Model-S provides a simple computational model of three observations that biologists have made, concerning the evolution of successful symbiotes.

Douglas (2010) provides an authoritative analysis and summary of research in symbiosis. Based on this work, we identified three prominent hypotheses about the characteristics of symbiotes that tend to be selected by evolution:

1. *Management:* Douglas (2010, p. 90) wrote, "… many symbioses are not associations of equals, but involve one organism that can control many of the traits of its partners." Also (Douglas, 2010, p. 72), "The incidence of cheating depends on opportunity, and opportunity can be minimized by the controlling partner." In this article, we call the controlling partner the *manager* and the other partners *workers*.

2. *Mutualism:* Douglas (2010, p. 56) wrote, "Conflict is inherent to the reciprocal exchange of benefits that underpin symbiosis." Furthermore (Douglas, 2010, p. 67), "Just as conflict arises from a difference in selective interest between the partners of a symbiosis, conflict can be resolved by increasing the overlap in selective interest of the partners." We call a partner that benefits from belonging to the symbiote an *insider*. A partner that does not benefit from belonging to the symbiote is an *outsider*. In Model-S, we can easily distinguish these two cases by measuring the growth of a partner both inside the symbiote and outside the symbiote. If the partner grows more inside, it is an insider; otherwise, it is an outsider. This kind of measurement is much easier in a computational model than in a biological organism, because it is usually difficult to remove a partner from a symbiote without harming it

3. *Interaction:* Douglas (2010, p. 58) wrote, "If repeated interactions with one partner reveal it to be a cooperator, then a player may have a vested interest in the continued well-being of that partner." We call a partner that interacts with the other partners an *ensemblist*. A partner that avoids interacting with the other partners is a *soloist*.

Model-S has two parts, a cellular automaton in which seed patterns grow and interact with each other and an external evolutionary algorithm in which seed patterns undergo mutation, crossover, asexual and sexual reproduction, and symbiosis (Turney, 2020; 2021a). The Golly software is used for the cellular automaton (Trevorrow et al., 2022) and Python code is used to implement the evolutionary algorithm (Turney, 2022). Both parts are freely available for downloading (see the two preceding references).

In Section 2, we introduce three variations on John Conway's Game of Life (Gardner, 1970). The first game, the original Game of Life, is usually displayed as a grid of two colours (black and white cells). The second game, the Immigration Game, created by Don Woods (Wainwright, 1971), is usually displayed as a grid of three colours (red, blue, and white cells). In this article, we introduce a new game, the Management Game, which is displayed as a grid of six colours (red, blue, orange, green, purple, and white cells). All three of these games follow the rules of the original Game of Life, except there are new rules governing the colours in the Immigration Game and the Management Game. The same patterns unfold in all three games, but the patterns are coloured differently. In particular, the new colours in the Management Game enable us to see interactions among groups of cells that cannot readily be seen in the Game of Life. The new colours are analogous to the practice of staining cells in biology experiments. The additional colours allow us to distinguish managers and workers, insiders and outsiders, and ensemblists and soloists.

Section 3 describes how Model-S works. Each run of Model-S last for 100 generations, where each generation involves 200 births, which yield 20,000 evolved organisms (symbiotes and individuals). Over a period of a month, we ran 40 instances of Model-S on three computers, resulting in a total of 800,000 organisms ($40 \times 20,000$). The fitness of the organisms varies greatly from the earlier generations of a run (relatively small and unfit organisms) to the later generations of a run (relatively large and fit organisms).

In the core of the paper, Section 4, we test the three hypotheses, *management*, *mutualism*, and *interaction*. Section 5 discusses limitations of our work and ideas for future work. We conclude in Section 6.

To explain some of the concepts involved in Model-S, it was necessary to introduce several technical terms. Table 2 provides a list of these words and their definitions, so that the reader can find the definitions without searching through the article. The definitions will be explained in more detail throughout the article.

Table 2. This table defines some technical terms that will appear frequently in this article. For ease of reference, the definitions are gathered together here. The Model-S terms are mostly borrowed from the domain of biology, whereas the other terms are common in the cellular automata community.

| Term | Definition | Domain |
|---|---|---|
| symbiote | a cellular automaton configuration composed of two or more partners, joined together, demarcated by a purple border line | Model-S |
| individual | a lone cellular automaton configuration, with no partners and no border lines | |
| organism | a general term that applies to both symbiotes and individuals | |
| partner | a member of a symbiote | |
| seed | the original state of an organism, before the cellular automaton game begins | |
| management | a symbiote with one manager and one or more workers | |
| mutualism | a symbiote with zero outsiders and two or more insiders | |
| interaction | a symbiote with zero soloists and two or more ensemblists | |
| genome | a seed may be considered as a genome, since the seed determines how an organism develops – interaction with other organisms also affects how the seed grows | |
| phenome | the phenome is the dynamic development of a seed over time, as the cellular automaton runs | |
| cell | a single square in the cellular automaton grid | Cellular Automata |
| parent | when three cells in a cellular automaton grid cause an empty cell to come alive, we say that the three cells are the parents of the new live cell | |
| player | the human player in a cellular automaton game, who designs the patterns (organisms) – the evolutionary algorithm in Model-S takes over the role of the human player | |

## 2 Cellular Automata: Three Games of Life

Figure 1 shows examples of the three cellular automata, the Game of Life, the Immigration Game, and the Management Game. The left column contains the initial seed patterns, the states of the games at time $t = 0$. The right column contains the final patterns, at time $t = 1000$. After 1000 time-steps, the games have usually settled into stable configurations, which are called *ashes* in the Game of Life community. Ashes are patterns that are static, oscillating, or moving through space (LifeWiki, 2022).

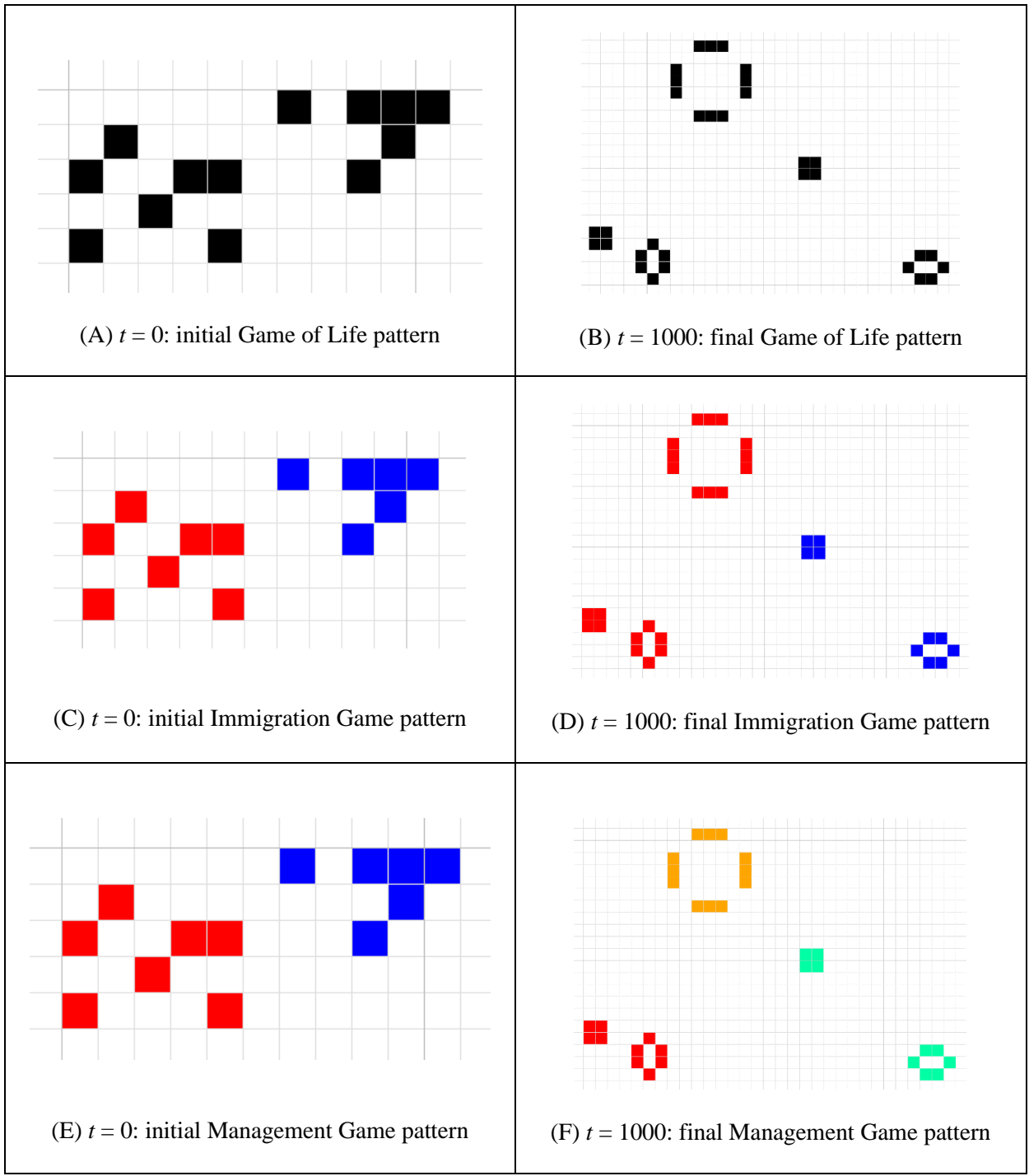

Figure 1. The Game of Life, the Immigration Game, and the Management Game at time $t = 0$ and $t = 1000$. Adding colours to the games reveals how parts of the initial patterns determine parts of the final patterns.

The Game of Life, the Immigration Game, and the Management Game are identical if we ignore colour, but colour gives us important information. In the Immigration Game, we can see that the blue pattern in Image C is responsible for creating the two blue patterns in Image D, and the red pattern in Image C is responsible for the three red patterns in Image D, but this information is not available in the Game of Life when we look at Images A and B. Likewise, the Management Game gives us information in Image F that is not visible in Image D.

## 2.1 The Game of Life

John Conway's Game of Life (Gardner, 1970) is played on an infinite, two-dimensional grid of square cells, where each cell is either *dead* (state 0) or *alive* (state 1). Often the dead state (the background) is coloured white and the live state (the foreground) is coloured black. The state of a cell changes with time, based on the state of its eight nearest neighbours (the *Moore neighbourhood*). Time passes in discrete intervals, and the states of the cells at time $t$ uniquely determine the states of the cells at time $t + 1$. There is only one player in the game and the player's only actions are to choose the initial states of the cells at time $t = 0$ and the time limit for the game, given by a maximum value for $t$. The states for $t > 0$ are calculated by a computer. The initial states form a seed pattern that determines the course of the game. The rules are summarized in Table 3.

Table 3. This table presents the rules of the Game of Life. Black cells are *alive* (state 1 in the cellular automata grid) and white cells are *dead* (state 0 in the grid). Let B represent *born*, S represent *survive*, D represent *die*, and U represent *unborn*. We use the convention that B3 implies its complement U01245678 and S23 implies its complement D0145678. This convention allows us to summarize the rules of the Game of Life as B3/S23.

| Description of change | State of 8 neighbouring cells at time $t$ | State of central cell at time $t$ | State of central cell at time $t + 1$ | Abbreviation |
|---|---|---|---|---|
| born | 3 black neighbours | white | black | B3 |
| survive | 2 or 3 black neighbours | black | black | S23 |
| die from overpopulation | 4 or more black neighbours | black | white | D45678 |
| die from underpopulation | 0 or 1 black neighbours | black | white | D01 |
| unborn | anything other than 3 black neighbours | white | white | U01245678 |

The rules for updating states in the Game of Life are compactly expressed as B3/S23: A cell is *born* (it switches from state 0 to state 1) if it has exactly three living neighbours (B3). A cell *survives* (it remains in state 1) if it has two or three living neighbours (S23). Otherwise, the cell *dies* (it switches from 1 to 0) or remains *unborn* (it remains in state 0).

## 2.2 The Immigration Game

Don Woods' Immigration Game (Wainwright, 1971) is almost the same as the Game of Life, except there are two different live states (states 1 and 2, usually represented by red and blue colours). The rules for updating remain B3/S23, but there are two new rules for determining colour: (1) Live cells do not change colour unless they die (they become white). (2) When a new cell is born, it takes the colour of the majority of its living neighbours. Since birth requires three live neighbours, there is always a clear majority. The initial states at time $t = 0$ are chosen by the two players of the game; one player makes a red seed pattern and the other player makes a blue seed pattern. The players agree on a time limit, given by a maximum value for $t$. The rules are summarized in Table 4.

Table 4. This table summarizes the rules of the Immigration Game. The Immigration Game follows the rules of the Game of Life, except when a new cell is born (B3); therefore, the table only shows the rules for birth. White cells are *dead* (state 0) and red and blue cells are *alive* (states 1 and 2). At time $t$, we have an empty central cell (state 0) with eight neighbouring cells around it. The empty central cell comes alive (it is born) at time $t + 1$ if exactly three of the neighbouring cells are alive at time $t$. This is the same as the Game of Life. The Immigration Game diverges from the Game of Life in the way that the colour of the central cell at time $t + 1$ is determined.

| State of 8 neighbouring cells at time $t$ | State of central cell at time $t$ (dead) | State of central cell at time $t + 1$ (alive) |
|---|---|---|
| 3 red neighbours | white | red |
| 3 blue neighbours | white | blue |
| 2 red neighbours + 1 blue neighbour | white | red |
| 2 blue neighbours + 1 red neighbour | white | blue |

In the Immigration Game, if states 1 and 2 were coloured black, instead of red and blue, the game would appear to be exactly the same as the Game of Life. The purpose of the two colours is to score the two players, to convert the Game of Life from a solitaire game into a two-player competitive game. In Model-S, we score each player by the *growth* of their initial seed, defined as the final number of cells of their colour minus the initial number of cells of their colour. Growth will be negative if the final number is less than the initial number. The motivation for this method of scoring is to avoid biasing the game in favour of the seed that has the most living cells at the beginning of the game. The winner is the player whose colour has grown the most over the course of the game.

## 2.3 The Management Game

The Management Game (a new game, introduced here) extends the Immigration Game by adding two more *live* states, orange and green. The Management Game also adds one more *dead* state, purple, to serve as a border for separating the partners of a symbiote. This brings the total number of states to six. Table 5 lists the six states and introduces some terminology.

Table 5. This table introduces some terminology that we will use to define the rules of the Management Game. In our experiments, the initial seeds at time $t = 0$ are always red or blue. Orange and green only appear later at $t > 0$ as red and blue interact with each other. The purple borders function as if they were dead, like the white background. The borders only appear at time $t = 0$. They turn white at time $t = 1$.

| States | Colours | Descriptions | |
|--------|---------|-------------|------|
| 0 | white | the background colour | dead |
| 1 | red | initial seed colour or a child born from 3 red parents | alive |
| 2 | blue | initial seed colour or a child born from 3 blue parents | alive |
| 3 | orange | red origins but with some past non-red contact | alive |
| 4 | green | blue origins but with some past non-blue contact | alive |
| 5 | purple | the colour of the borders that separate partners in a seed | dead |
| 2, 3, 4 | non-red | blue, orange, or green | alive |
| 1, 3, 4 | non-blue | red, orange, or green | alive |
| 2, 4 | blue/green | blue or green | alive |
| 1, 3 | red/orange | red or orange | alive |

Table 6 presents the rules of the Management Game. Like the Game of Life and the Immigration Game, the basic rule is still B3/S23. The only change is how colours are handled. We colour one partner of the symbiote red and the other partners blue and then play the Management Game. If red and blue do not interact, then we will only see changing red and blue patterns. If red and blue do interact, then we will start to see orange and green colours appearing.

Table 6. This is a summary of the rules of the Management Game. We focus here on the rules for the birth of a new cell (B3), since this is the only case where the Management Game differs from the Game of Life and the Immigration Game. At time $t$, we have an empty (white or purple) cell with eight neighbouring cells around it. For birth to happen in the empty central cell, exactly three of the eight neighbours must be alive. The table specifies the colour the central cell will have at time $t + 1$, based on the colours of the neighbouring cells at time $t$.

| State of 8 neighbouring cells at time $t$ | State of central cell at time $t$ (dead) | State of central cell at time $t + 1$ (alive) |
|---|---|---|
| 3 red neighbours | white or purple | red |
| 3 blue neighbours | white or purple | blue |
| 2 red/orange + 1 non-red neighbour | white or purple | orange |
| 2 blue/green + 1 non-blue neighbour | white or purple | green |

We noted above that, if states 1 and 2 in the Immigration Game were displayed as black, then the Immigration Game would look exactly like the Game of Life. The only difference between the Game of Life and the Immigration Game is colouration. Likewise, in the Management Game, if state 3 (orange) were displayed as red, state 4 (green) were displayed as blue, and state 5 (purple) were displayed as white, then the Management Game would look exactly like the Immigration Game. In the Management Game, orange is used to mark a subset of the cells that would be red in the Immigration Game, green is used to mark a subset of the cells that would be blue, and purple is used to mark a subset of the cells that would be white.

# 3 Model-S: A Model of the Evolution of Symbiosis

Model-S is a genetic algorithm for evolving seed patterns that are good at playing the Immigration Game (Turney, 2020; 2021a; 2022). It is based on a GENITOR-style genetic algorithm (Whitley, 1989), with one-at-a-time reproduction, a constant population size, and rank-based tournament selection. Model-S uses Golly (Trevorrow et al., 2022), an open source, cross-platform application for exploring cellular automata. Golly can be controlled with Python code. Model-S consists of Python routines that manage a population of seed patterns, passing the patterns on to Golly in order to measure their fitness.

Model-S uses the Immigration Game as a contest for measuring the fitness of seed patterns. Two black and white seed patterns are sampled from the population. One of the seeds is altered by switching black cells to red and the other seed is altered by switching black cells to blue. These two seeds are the initial states of the two players of the Immigration Game. Model-S selects a time limit for playing the game, based on the sum of the live cells in the two seeds. Larger seeds are given more time, since it takes longer for them to settle into stable configurations (*ashes*). The growth of the red seed is the total number of live red cells when the time limit is reached minus the number of live red cells in the initial seed pattern. The growth of the blue seed is measured likewise. The winner of the Immigration Game is the seed that grows the most. A given seed's fitness is measured by the average number of Immigration Games it wins when competing against every seed in the current population.

It would have been simpler to measure the fitness of a seed alone, rather than using one-on-one competitions, but we believe that competition is a core aspect of biological evolution by natural selection. Biological organisms do not evolve alone.

## 3.1 Time Scales

There are three different scales of time involved in Model-S. First, each competition between two seeds takes a certain amount of time, depending on the size of the two competing seed patterns. This time is spent running the Immigration Game in the Golly cellular automaton software (*game time*). Second, each seed has a *lifetime*, which begins when it enters the population and ends when it leaves the population. This time passes in the Python code that implements the model of evolution. Third, there is the length of time that a particular instance of Model-S runs, from the first generation to the last generation. This time also passes in the Python code (*evolutionary time*).

A typical Immigration Game runs for about a thousand steps (*game time*). Model-S has a constant population size of 200. The initial population consists of 200 randomly generated entities. The initial seed patterns are small (5×5 cell grids) and none of them are symbiotes. Each subsequent newly born entity replaces an existing entity in the population (the least fit member of the population is replaced). When 200 new entities have been born, we say that one generation of time has passed. Model-S runs for 100 generations, which results in 20,000 evolved seeds (200×100). The initial 200 random seeds are considered as generation zero; they are not considered as evolved seeds. We store all of the seeds for later analysis.

## 3.2 Four Layers

Model-S is constructed with four layers, each subsequent layer building on the previous layers, as shown in Figure 2. The purpose of having four layers is to measure each layer's contribution to the fitness of the evolving population, by selectively enabling or disabling layers. In the current article, we use all four layers.

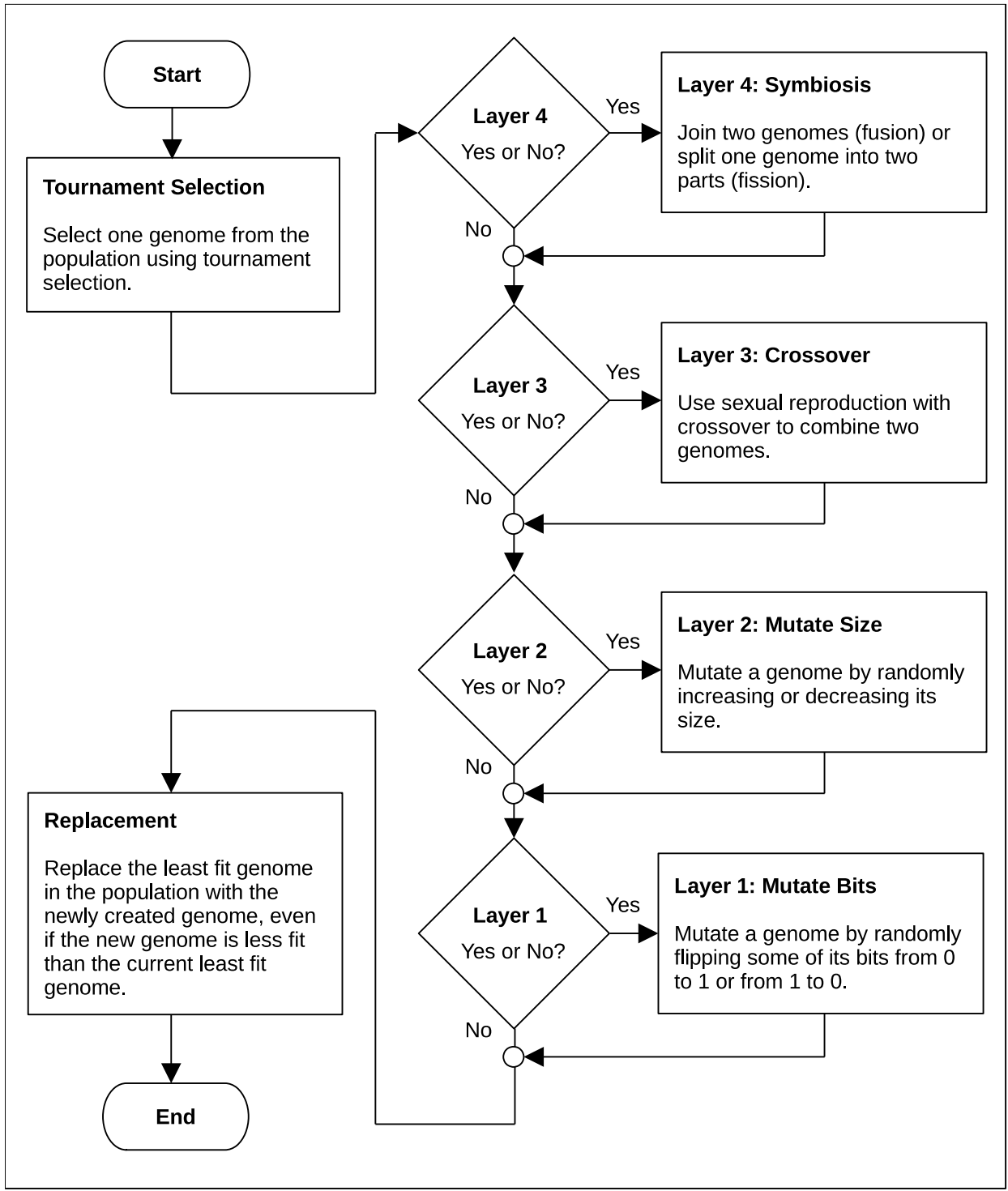

Figure 2. This flowchart outlines the process for selecting an individual's genome from the population and creating a new genome. This process is a subroutine in a loop that produces a series of new individuals. The decision to use a given layer is determined by the parameters of Model-S. This flowchart first appeared in a previous article (Turney, 2021a). A more detailed four-figure flowchart is also available (Turney, 2020).

Layer 1 implements a simple form of asexual reproduction, with a fixed genome size (that is, a fixed matrix size). A member of the population is selected for reproduction using tournament selection (Whitley, 1989). The chosen seed pattern is mutated by randomly flipping some of the bits in the seed matrix and it then competes in a series of one-on-one Immigration Games with the other members of the population. Its fitness is the average fraction of games it wins.

Layer 2 implements a slightly more sophisticated form of asexual reproduction. A member of the population is selected for reproduction using tournament selection. Layer 2 allows the seed matrix to grow or shrink by appending or removing a row or column to or from the seed matrix. Layer 2 then passes the seed on to Layer 1 for mutation by flipping bits.

Layer 3 selects two seeds from the population using tournament selection and then combines them with genetic crossover (sexual reproduction). Layer 3 requires the two seeds to be somewhat similar for crossover to proceed. If a suitable match is found, then Layer 3 combines the two seeds with crossover and passes the new seed on to Layer 2 for row or column adjustments. Otherwise, if no suitable match is found, Layer 3 passes only one of the two seeds on to Layer 2, without making any changes to the chosen seed.

Layer 4 adds symbiosis to Model-S. Two seeds are selected from the population and they are fused together, side-by-side, creating a new symbiotic genome. This new fused seed is treated as a whole; that is, selection shifts from the level of the two partners to the level of the whole. The parameters in Model-S are set so that fusion is rare (the probability of fusion is set to 1 chance in 200; that is, an average of one new fusion in each generation). Most of the time, Layer 4 makes no changes and simply passes control to Layer 3. The main result of our past work (Turney, 2020) is evidence that symbiosis (Layer 4) promotes fitness improvements in the Immigration Game.

Each symbiote created by Layer 4 is composed of partners (seed patterns) that have been fused together to make a new whole (a new, fused seed pattern). In Layer 4, seed patterns are fused two at a time. In the earliest generations of a run of Model-S, when two seeds are selected for fusion, they will not have experienced fusion before; therefore, the selected seeds will not be composed of partners. The result of fusion with these seeds will be a new seed with two partners. Over the course of a long run of Model-S, eventually one or both of the two seeds that are selected for fusion will have experienced fusion in the past, so the new, fused seed will have three or more partners. In our 40 runs of Model-S, there are seeds with two, three, four, and five partners.

The state of a seed pattern at time $t = 0$ is analogous to the genome of an organism. When a seed pattern enters a game, it develops over time, following the rules of the game, as $t$ increases. The developing pattern is analogous to the developing phenome of an organism. The genome is static and the phenome is dynamic.

Model-S records the genome of each organism (the pattern at time $t = 0$), but it keeps no record of the phenome (any patterns at $t > 0$). All the information that Model-S requires for evolving new genomes is provided by the fitness score of the phenome. The phenome itself is ephemeral.

When Model-S is running, symbiotes and individuals compete against other symbiotes and individuals, in order to measure the fitness of the entities in the population. The competitions use the Immigration Game to compare two entities sampled from the population. These competitions determine which entities can reproduce; that is, the competitions determine life and death in the population. All entities, symbiotic or not, are treated the same way. Model-S does not analyze the partners inside a symbiote in any way: a symbiote is treated as a whole.

### 3.3 Measuring Fitness

Fitness in Model-S is measured by pairwise contests, in which two seeds compete, each attempting to grow more than the other. Therefore, fitness is relative to the current population; there is no absolute measure of fitness built into Model-S. The average fitness of a given population is always 50%. Every win in the population is counterbalanced by a loss. This means that the fitness of a seed in one population tells us nothing about how fit that seed would be in another population. This is like real life, where fitness is always relative to the given population.

However, we store a copy of every seed so that we can study seeds and experiment with them after Model-S is done running. Hence, we can sample seeds from early generations and make them compete against seeds from the later generations, and we find that the later seeds typically win when competing against the earlier seeds. We can use comparisons across different generations and across different runs of Model-S to rank seeds in a relative way. This allows us to determine whether there are some characteristics that lead to higher levels of fitness in a population.

In this section, we will look at fitness in general, without considering management, mutualism, and interaction. In Section 4, we will examine whether management, mutualism, and interaction are characteristics of symbiotes with greater fitness.

The fitness of a seed at any point in time is the fraction of contests it wins in competitions with the other seeds in its population. A contest begins with two seeds randomly placed on a toroid (the surface of a doughnut). It is more common in the Game of Life community to use an unbounded plane as the playing surface, but we chose to use a toroid, because a toroid forces the two seeds to interact more than they would on an unbounded plane. The size of the toroid is proportional to the size of the seeds: larger seeds are given more space to grow. The length of time allowed for a contest also increases as the seeds become larger.

To score a contest between two seeds, it is sufficient to use the Immigration Game, with red and blue players on a white background, but it is also possible to use the Management Game, which can give us more insight into what is happening in the game. To calculate the score with the Management Game, we simply treat orange as if it were red, and we treat green as if it were blue (see Figure 1).

Figure 3 shows a contest between two seeds sampled from generation 20 of a run of Model-S. The upper grid shows the initial seeds, at the beginning of the contest, and the lower grid shows the final results. Note that the two seeds in the upper grid are individuals, not symbiotes: there are no purple borders demarcating partners. In the final pattern, we see that the red-and-orange side has won against the blue-and-green side.

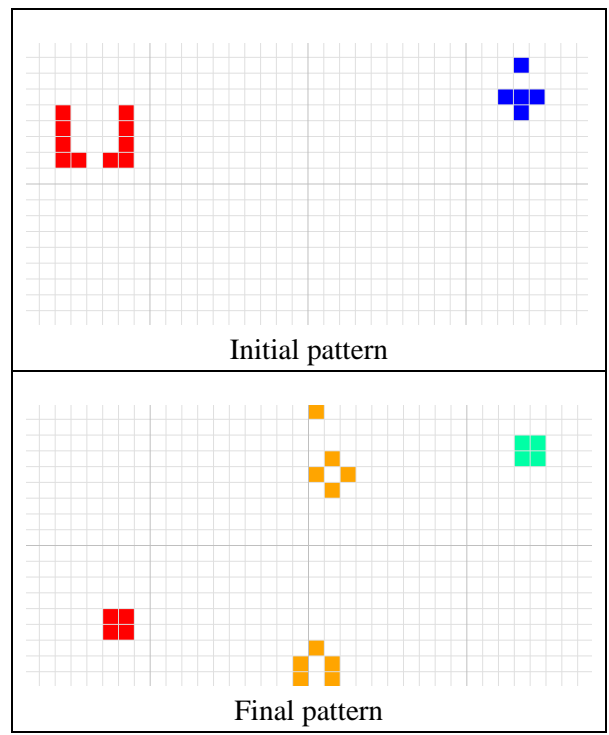

Figure 3. Here we see two individuals fighting for survival, red seed versus blue seed. The two rectangular grids above represent one toroid at two different points in time. The top of a grid wraps around to join the bottom and the left side wraps around to join the right side. The orange square at the top of the lower grid is thus directly connected to the orange object at the bottom of the lower grid.

In general, the population in early generations of a run of Model-S consists mostly of individuals whereas the population in later generations consists mostly of symbiotes. Generation zero consists entirely of randomly generated individuals. The evolutionary algorithm in Model-S uses these random individuals as the starting point for mutation and selection, to produce the next generations.

Figure 4 presents a contest between two seeds sampled from generation 90 of a run of Model-S. Again, the red-and-orange side has won against the blue-and-green side. The purple borders show that these seeds are symbiotes, not individuals. Both symbiotes are composed of three partners. We can see that the two seeds in Figure 4 have similar patterns of live cells, with a few small differences. These little mutations slowly add up to significant differences in fitness. The outcomes of these competitions are sensitive to the relative locations and orientations of the two players, but the large number of competitions smooths out the noisy signal, so that evolution can take place. With a population of 200, each seed must face ongoing competitions with 199 other seeds.

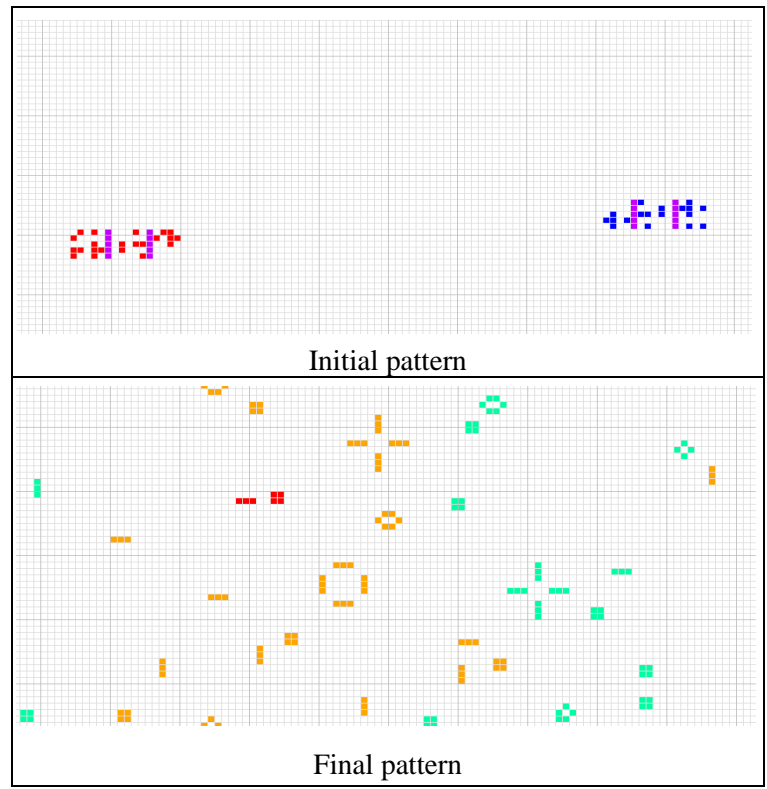

Figure 4. Here we see two symbiotes competing. Note that their toroids are larger, in proportion to the size of their larger seeds, compared to the seeds in Figure 3. Both symbiotes contain three partners.

Note that seeds (both individual seeds and symbiotic seeds) compete one-on-one to determine their fitness, but there is no incentive for competition among the partners inside a symbiote. Symbiotes survive or perish as a unit. Internal competition reduces their ability to compete with other organisms. To maximize fitness, symbiotes should be composed of partners that cooperate well internally, but compete strongly against other symbiotes. Many symbiotes may have internal competition, especially in the early generations of a run, but evolution tends to reduce such internal competition, as we will see in Section 4.

Figure 4 reveals the growth of organisms in their lifetimes, from birth as initial seed patterns to maturity as final patterns, whereas Figure 5 plots the growth of seeds over generations, from parent to child to grandchild. Seeds are the genomes of the organisms. The genome determines how an organism will grow and how it will compete with other organisms. To evolve complex and adaptive behaviours, seeds must evolve to larger sizes, in order to encode more information for increasingly complex winning strategies.

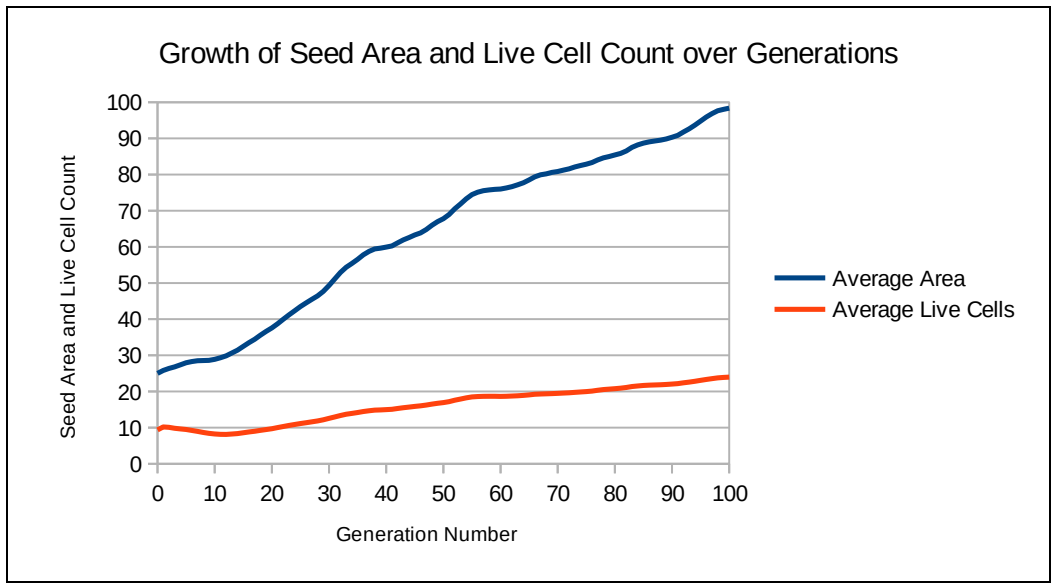

Figure 5. This graph shows that seeds increase in area and in the number of live cells as the generations pass.

In Figure 5, we see that the area of the seeds increases faster than the number of live cells in the seeds. Fitness is based on the fraction of games won and winning is based on which seed grows the most in one-on-one competition. The growth of a seed is the final number of live cells minus the initial number of live cells. Our reasoning for subtracting the initial cells is that we did not want to give an unfair advantage to seeds that begin their lives with a large number of live cells.

Figure 5 shows that the average *number* of live cells in a seed slowly increases over the generations, from 10 to 24, whereas the average *area* of a seed increases more rapidly, from 25 to 98. The density (live cells per area) is thus 0.40 in the first generation and 0.24 in the final generation. This downward trend in density may be a side-effect of inserting purple borders between partners in a symbiote, since the borders create gaps between the partners. Consider the red seed in Figure 4, which has an area of $16 \times 5 = 80$. If we delete the two purple borders and the empty column, the area is then $13 \times 5 = 65$. Clearly the seeds are not packed as tightly as they could be, but we do not see this as a problem.

Over generations, the seeds also increase in the number of partners they contain, as shown in Figure 6. The population begins with single individuals. Symbiotes with two partners soon appear. The final population mostly contains symbiotes with two to four partners. It appears that the upward trend will continue indefinitely. Unfortunately, the simulation runs much more slowly as the symbiotes grow larger, which means that significantly extending the number of generations in Model-S will require more patience and more computational power.

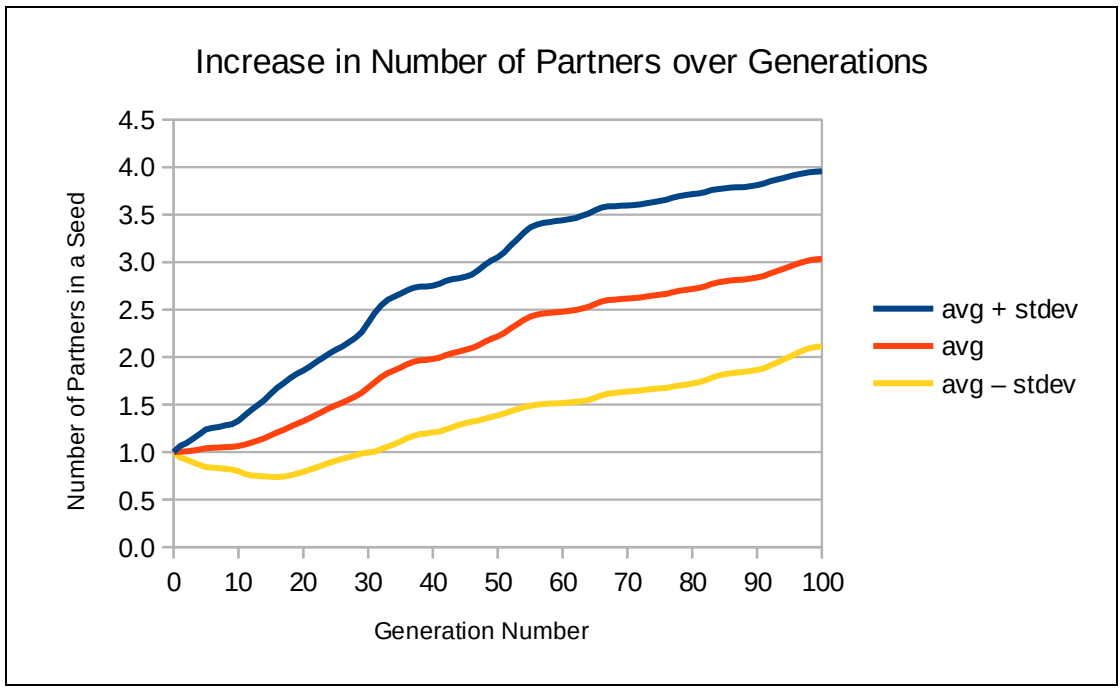

Figure 6. The initial generation consists entirely of individuals, but symbiotes of two or more partners soon appear. In the final generation, an average symbiote has three partners, but two or four partners is not uncommon.

Seeds do not merely increase in size over generations. Fitness is measured by one-on-one competitions, not by size. The increase in size is a necessary side-effect of increasingly complex structures that have evolved to win competitions. A small seed can only encode a small amount of information. Increasing fitness requires increasing information for winning strategies, which requires increasingly larger seeds to encode the strategies.

To demonstrate this, we can make seeds compete with shuffled versions of themselves. We take a seed from the population, randomly choose two cells in the seed, and then swap the values (alive or dead) in the two cells. The swapping is repeated many times. This procedure creates a new shuffled seed that has exactly the same area and exactly the same number of live cells as the original seed. The only difference between the two seeds is that the shuffled seed is thoroughly randomized, whereas the original seed has a structure that has evolved to perform well in one-on-one competitions for growth.

In Figure 7, we see what happens when evolved seeds compete against shuffled versions of themselves. In generation zero, when evolution has not yet begun, the fraction of games won by the seeds is 50%, because the seeds in generation zero have not yet undergone evolution. Their fitness is the same as the fitness of their shuffled opponents. However, as the generations pass, the evolved seeds increasingly outperform their shuffled counterparts. In the final generation, the evolved seeds win 95% of the games. While size is likely an advantage when two seeds compete, it is clear that size (measured in area or number of live cells) is not enough; evolved structure is the key to success.

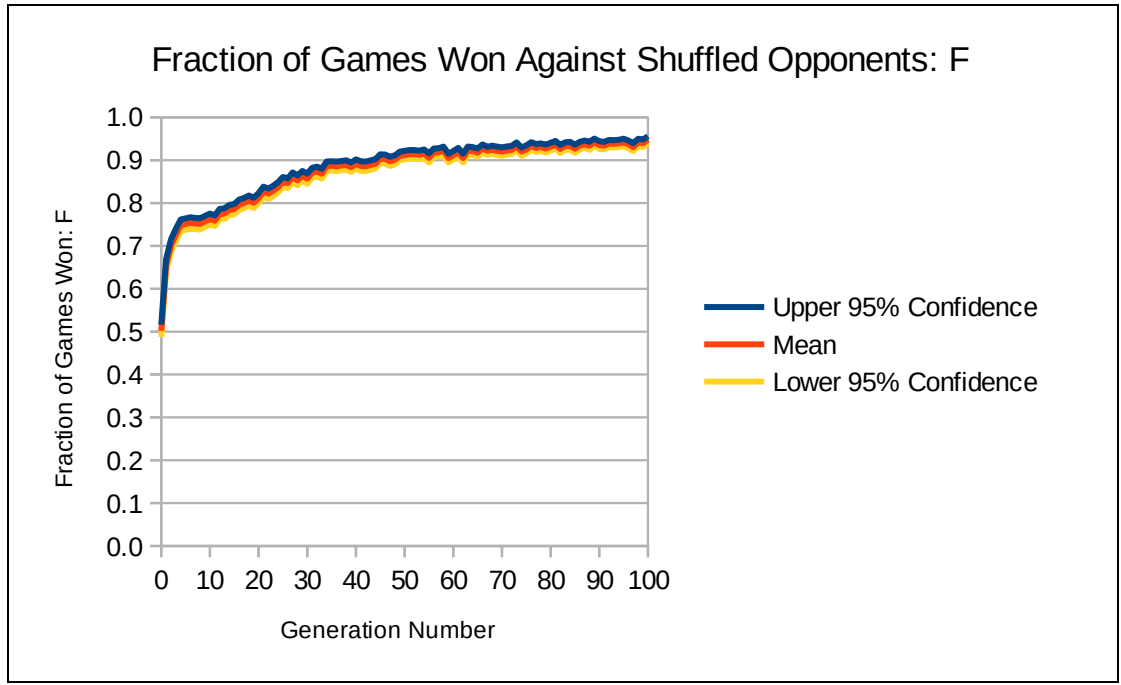

Figure 7. As evolution proceeds, evolved seeds increasingly outperform their shuffled counterparts. The data used here is from 40 runs of Model-S, so the 95% confidence intervals around the mean are tight (that is, the confidence is high).

By definition, the fraction of games won cannot pass 1.0. Looking at Figure 7 suggests that progress is flattening out and evolution in Model-S is stalling, but this flattening is a consequence of measuring success by the fraction of games won. Let $F$ be the fraction of games won. If we replace $F$ with the formula $1 / (1 - F)$, then we can see in Figure 8 that progress is continuous. The trend in Figure 8 is approximately linear (with some noise).

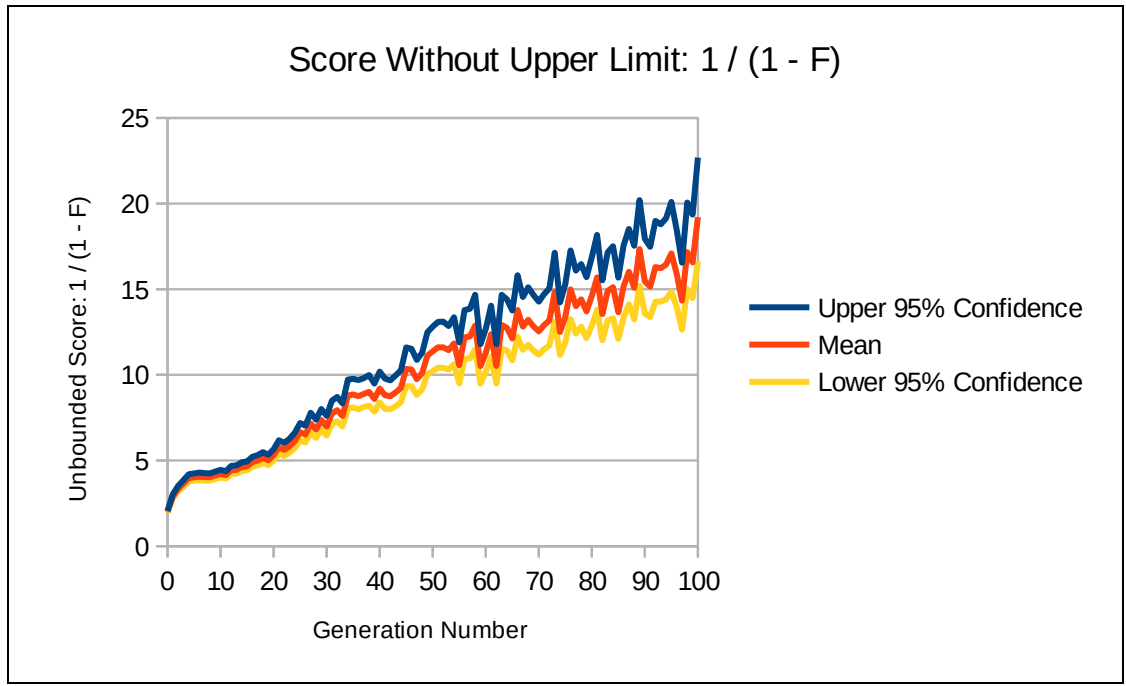

Figure 8. This plot provides another way of viewing evolutionary progress in Model-S. The formula eliminates the fixed upper bound of 1.0 that is built into measuring improvement by the fraction of games won.

## 4 Three Hypotheses about the Evolution of Symbiotes

The Immigration Game is sufficient for evolution in Model-S. It is a two-person game with a clear goal: the winner is the player that grows the most. The Management Game is not necessary in Model-S. However, after we have run 40 instances of Model-S and collected the data from these runs, the Management Game becomes a useful tool for data analysis, outside of Model-S. The extra colours in the Management Game allow us to see interactions that are not visible in the Immigration Game.

Figure 9 shows how we analyze the partners in a three-partner symbiote. Given the initial seed pattern of the symbiote, we focus on one of the partners in the symbiote and colour that partner red. The other partners are coloured blue. We then run the Management Game for 1000 steps and observe the results. This process is repeated for each of the three partners in the symbiote.

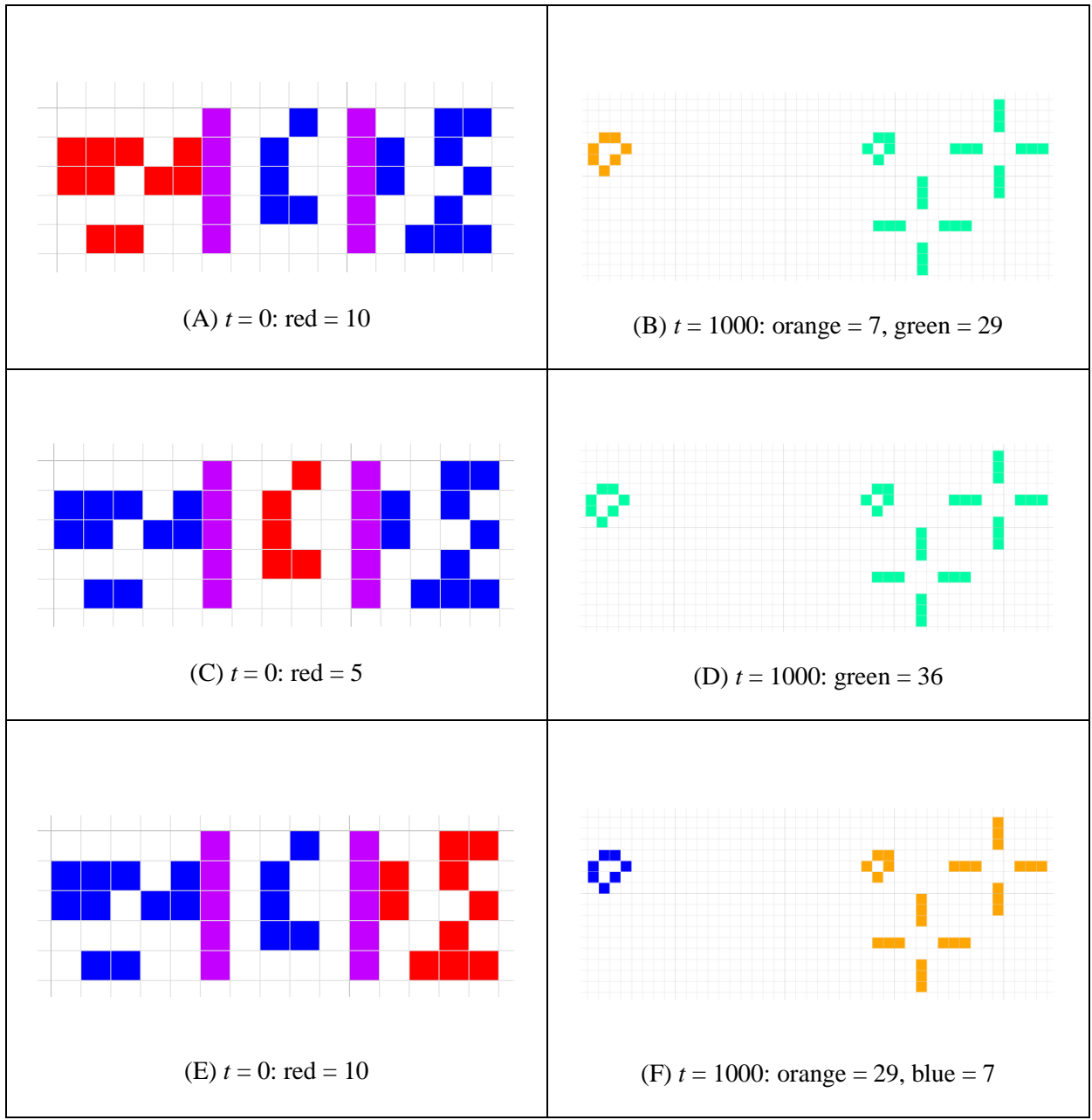

(A) $t = 0$: red = 10

(B) $t = 1000$: orange = 7, green = 29

(C) $t = 0$: red = 5

(D) $t = 1000$: green = 36

(E) $t = 0$: red = 10

(F) $t = 1000$: orange = 29, blue = 7

Figure 9. Image A shows an initial seed pattern ($t = 0$) with three partners, one red partner and two blue partners. The three partners are separated by purple borders, which disappear when $t > 0$. We focus on the red partner and watch how it interacts with the two blue partners. Image B shows the final pattern after one thousand steps ($t = 1000$) in the Management Game. By shifting the red partner from A to C to D, we can see how the red partner affects each of the final patterns, from B to D to F.

Given the seed pattern in image A, we can tell that the red partner played a majority role in creating the orange pattern in image B, and it played a minority role in creating the green patterns in image B. This follows from the rules of the Management Game (see Table 6). In image C, the central red partner played a minority role in all of the green patterns in image D. In image E, the rightmost red partner played a majority role in creating the orange patterns in image F. The rightmost red partner had no role in creating the blue pattern in image F. The blue pattern in image F was created solely by the two blue patterns in image E.

The four rows of Table 6 tell us four things: (1) A newly born red cell is always the work of three other red cells. (2) A newly born blue cell is always the work of three other blue cells. (3) When we see an orange cell, we know that, at some time in the past, a blue or green cell played a *supporting* role (one of three cells for B3) in forming the orange cell (since the initial seed pattern contains no orange). The *leading* role (two of three cells) was played by orange or red. (4) When we see a green cell, we know that, at some time in the past or in the current birth, a red or orange cell played a *supporting* role (one of three cells for B3) in forming the green cell (since the initial seed contains no green). The *leading* role (two of three cells) was played by green or blue.

## 4.1 Management

Suppose we have a symbiote at time $t = 0$ with $N$ partners. Let us focus on one of those partners and colour that partner red. All other partners are coloured blue. Then we run the Management Game for $t = 1000$ steps and observe the results. If the number of orange cells is greater than the number of green cells at $t = 1000$, then we define the red partner of the seed at $t = 0$ as a *manager*. If the number of orange cells is less than or equal to the number of green cells, then we define the red partner as a *worker*.

The motivation for these definitions of *manager* and *worker* is the observation that, when orange dominates green, we know the red partner played a leading role in creating the orange cells (see Table 6). Furthermore, the red partner did not work alone. If it had worked alone, the final pattern it created would be red, not orange. Thus, the orange cells represent both a leading role and interaction with others partners of the symbiote. On the other hand, green cells represent both a supporting role and interaction with other partners of the symbiote.

Looking at Figure 9, images A and B, we can see that the leftmost partner of the symbiote is a worker (because green dominates orange). Images C and D tell us that the central partner of the symbiote is also a worker (because green dominates orange). Images E and F indicate that the rightmost partner of the symbiote is a manager (because orange dominates green). Thus, the three-partner seed in Figure 9 consists of one manager and two workers.

In this article, we say that *management* happens when a symbiote has one manager and one or more workers. We expect that symbiotes containing a large number of partners will need more than one manager, but our experiments so far have only involved a few partners. We will leave for future work the issue of finding a general formula for the best ratio of managers to workers.

Our definition of *management* is directly based on the observations of biologists who study symbiosis. As Douglas (2010, p. 90) wrote, "… many symbioses are not associations of equals, but involve one organism that can control many of the traits of its partners." One partner (the manager) controls and the other partners (the workers) follow.

## 4.2 Mutualism

We define an *insider* as a partner inside a symbiote that benefits from being inside the symbiote, whereas an *outsider* is a partner inside a symbiote that would benefit from leaving the symbiote. When there is mutualism in a symbiote, all of the partners are insiders; all of the partners benefit from being in the symbiote. We use growth as a measure of benefit. If a partner grows more when it is inside the symbiote than when it is outside the symbiote, then the partner benefits from being in the symbiote: it is an *insider*. Otherwise, the partner is an *outsider*.

In biology, it is often not possible to remove a partner from a symbiote without damaging the partner. In Model-S, it is easy to remove a partner without damaging it. A computational model of mutualism allows us to perform experiments that would be very difficult with living creatures.

Figure 10 shows what happens when we extract the three red partners from the symbiote in Figure 9 and then allow them to grow individually, outside of the symbiote. Their growth as individuals (Figure 10, right column) is quite different from their growth in the symbiote (Figure 9, right column).

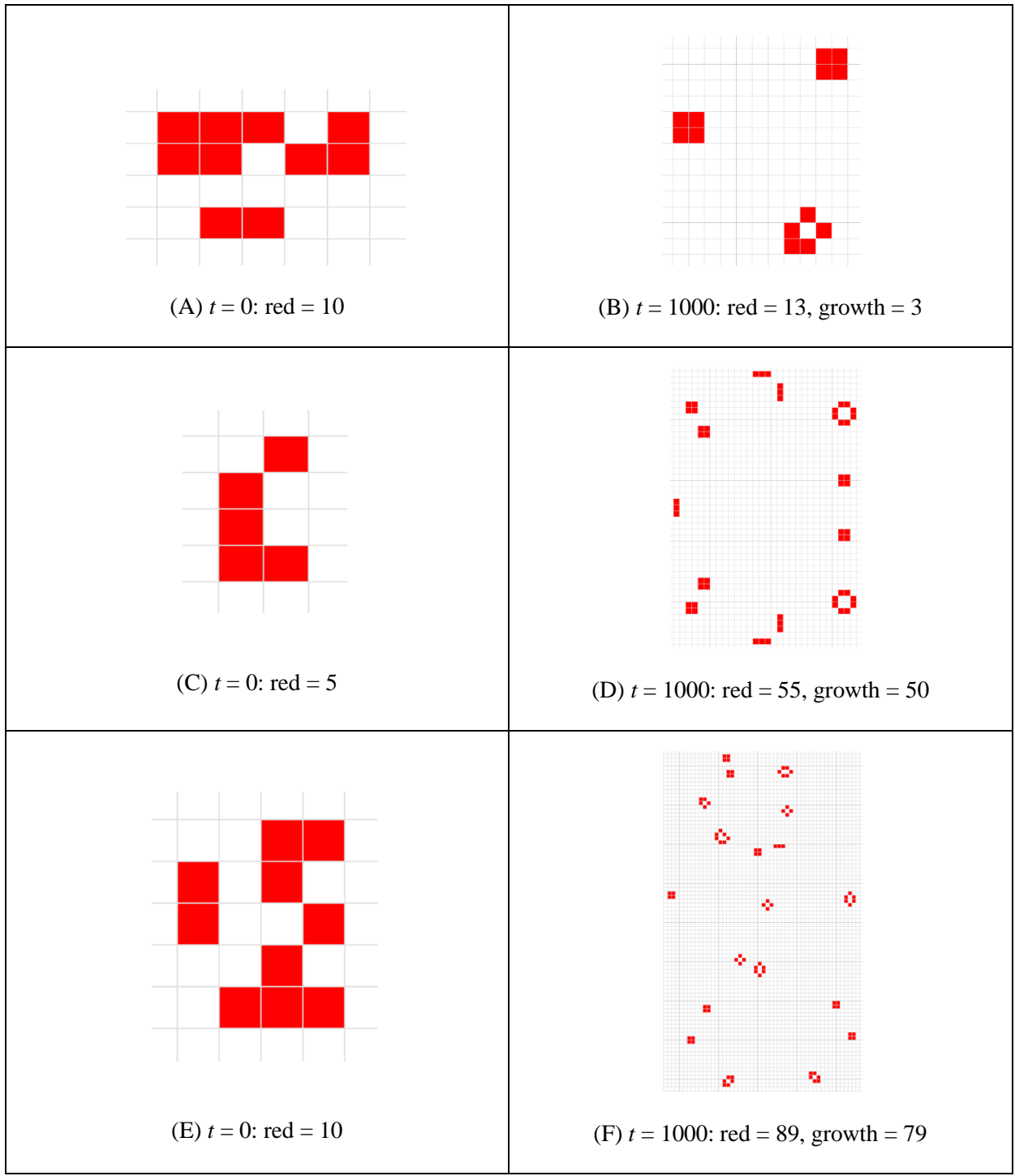

Figure 10. The three red patterns in the left column here (A, C, E) correspond to the three red patterns in the left column of Figure 9, but here they have been separated from their blue partners. Comparing Figure 9 and Figure 10, we can see how the growth of a partner in a symbiote differs from the growth of the same partner when alone.

We can easily measure the growth of the individuals (the partners removed from the symbiote), as shown in Figure 10: The first partner grows by 3 cells (B in Figure 10), the second partner grows by 50 cells (D), and the third partner grows by 79 cells (F).

Measuring the growth of the partners inside the symbiote (Figure 9) is more complicated, because we need to take the various colours into account. We weight the colours as follows: red = 1, blue = 0, orange = 2/3, green = 1/3. The reasoning here is that the initial red seed should get full credit for any new red cells that it creates (that is, each new red cell has a value of 1). The initial red seed should not get any credit for blue cells, since it cannot create any blue cells (see Table 6: only blue cells can make new blue cells). The initial red seed should get a credit of 2/3 for each new orange cell that it creates, since an orange cell is the work of two red or orange cells (and one other cell that is neither red nor orange). The initial red seed should get a credit of 1/3 for each new green cell that it creates, since one red or orange cell (and two other cells that were neither red nor orange) was responsible (at some time in the past or present) for creating the new green cell.

Table 7 gives an example of how to measure the growth of partners inside a symbiote (as in Figure 9) and outside a symbiote (as in Figure 10). We can see from the table that seed A grows more when it is inside the symbiote (4.33 inside versus 3 outside), whereas seed C grows more when it is outside the symbiote (7 inside versus 50 outside) and D also grows more when it is outside than inside (9.33 inside versus 79 outside). Thus, the three-partner symbiote in Figure 9 consists of one insider and two outsiders.

Table 7. This table shows how the growth of the partners in a symbiote is calculated, using the symbiote in Figure 9 and its partners in Figure 10 as an example. The growth of A into B (4.33) is greater in Figure 9, when the partners are together in the symbiote, than in Figure 10 (3.00), when the partners are alone. Therefore A, the first red partner in Figure 9 and Figure 10, is classified as an *insider*. The other two partners (C and E) are classified as *outsiders*, since they grow more outside of the symbiote.

|  |  | Red | Blue | Orange | Green | Weighted total |
|---|---|---|---|---|---|---|
|  | Colours: | Red | Blue | Orange | Green | Weighted total |
|  | Weights: | 1.000 | 0.000 | 0.666 | 0.333 | |
| Growth together in Figure 9 | A → B | -10 | -15 | 7 | 29 | 4.33 |
|  | C → D | -5 | -20 | 0 | 36 | 7.00 |
|  | E → F | -10 | -8 | 29 | 0 | 9.33 |
| Growth apart in Figure 10 | A → B | 3 | 0 | 0 | 0 | 3.00 |
|  | C → D | 50 | 0 | 0 | 0 | 50.00 |
|  | E → F | 79 | 0 | 0 | 0 | 79.00 |

We say that *mutualism* happens when a symbiote has zero outsiders and two or more insiders. That is, if every partner benefits from being in the symbiote, then there is mutualism. As in the case of management, when symbiotes contain a large number of partners, it may make sense to allow a small number of outsiders as exceptions to the rule, but we leave this question for future work.

Our definition of *mutualism* is also based on the observations of biologists. As Douglas (2010, p. 67) wrote, "Just as conflict arises from a difference in selective interest between the partners of a symbiosis, conflict can be resolved by increasing the overlap in selective interest of the partners." When all partners are insiders, all partners benefit from being part of the symbiote. A partner that thrives outside of the symbiote but languishes inside the symbiote is not a good partner.

## 4.3 Interaction

We define a *soloist* as a partner in a symbiote that prefers to work on its own, whereas an *ensemblist* prefers to work as partner in a team. A soloist avoids interaction with others, but an ensemblist seeks interaction. Given a red partner in a symbiote, if the red partner produces more red cells, then it is growing, but it is not interacting with the other partners of the symbiote. If the red partner produces orange or green cells, then it is interacting with the other partners of the symbiote. If the sum of orange growth and green growth is greater than the red growth, then we say that the partner is an *ensemblist*. Otherwise, if red growth dominates orange and green growth, then we say that the partner is a *soloist*. Blue growth is irrelevant, since the red partner cannot cause a new blue cell to appear (see Table 6). Only blue cells can make more blue cells.

Looking at Figure 9, we see that all three partners (the red partners in A, C, and E) produce some orange or green growth (in B, D, and F), but none of them produce red growth. Thus, the symbiote in Figure 9 has three ensemblists and zero soloists.

We say that *interaction* happens when a symbiote has zero soloists and two or more ensemblists. As with management and mutualism, we expect this rule will need to be relaxed for symbiotes that have large numbers of partners.

Our definition of *interaction* also comes from observations of biologists. As Douglas (2010, p. 58) wrote, "If repeated interactions with one partner reveal it to be a cooperator, then a player may have a vested interest in the continued well-being of that partner." Soloists who work alone are not contributing as much to the symbiote as ensemblists who interact with their partners.

## 4.4 Evolution with Management, Mutualism, and Interaction

We present three hypotheses: (1) Evolution will select for *management*: one manager and one or more workers. (2) Evolution will select for *mutualism*: zero outsiders and two or more insiders. (3) Evolution will select for *interaction*: zero soloists and two or more ensemblists. Over the generations, as evolution proceeds in Model-S, the number of cases of management, mutualism, and interaction will increase, and the increase will be faster than the increase in symbiotes in general. (By *symbiotes in general*, we mean the set of all symbiotes, including those with or without management, mutualism, or interaction.)

Figure 11 shows five groups of organisms, consisting of various subsets of the population, for each generation. There is no overlap between the set of individuals and the set of symbiotes: they are mutually exclusive and their union is the whole population for a given generation. The other three sets (management, mutualism, and interaction) are all subsets of the set of symbiotes, and all three sets intersect each other in varying amounts.

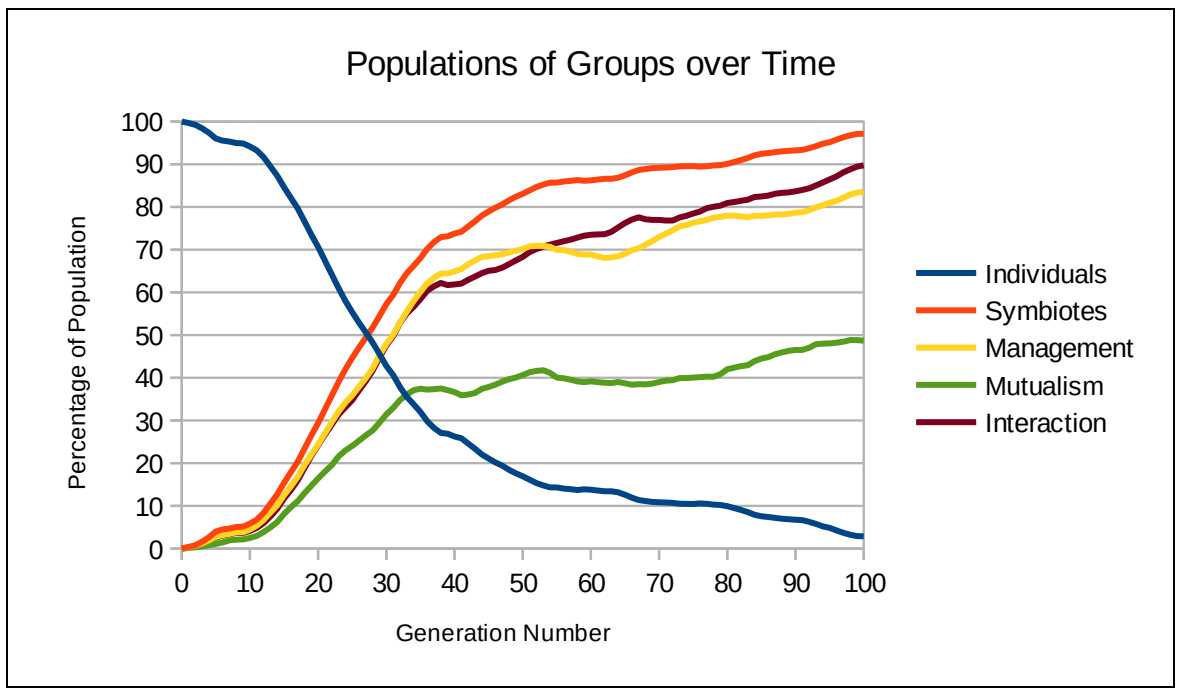

Figure 11. This graph plots five different groups as percentages of the population over the generations. The populations of management, mutualism, and interaction increase over the generations. The total population in any generation of one run is always 200. This graph is based on 40 runs, which yields $40 \times 200 = 8,000$ numbers for each point on the graph.

Since the population is fixed at 200 organisms, any increase in symbiotes (the red curve in Figure 11) must be matched by a decrease in individuals (the blue curve in Figure 11). Naturally the red and blue curves intersect at 50%. By generation 100, few individuals remain in the population: symbiotes dominate.

The populations of symbiotes with management, mutualism, and interaction increase over the generations, necessarily bounded by the increasing population of symbiotes in general. The upward trend in management, mutualism, and interaction suggests that evolution is selecting for these three properties, but it might be that "A rising tide lifts all boats." That is, the general rise in the population of symbiotes might be what passively causes the rise in management, mutualism, and interaction.

To make a case that evolution is selecting for management, mutualism, and interaction, we need to show that these three groups are rising faster than the group of symbiotes in general. We need to show that the increase in management, mutualism, and interaction is driven, not passive. Figure 12 presents the growth of the three groups relative to the growth of symbiotes in general. We plot the growth of each of the three groups as a percentage of the growth of the group of all symbiotes.

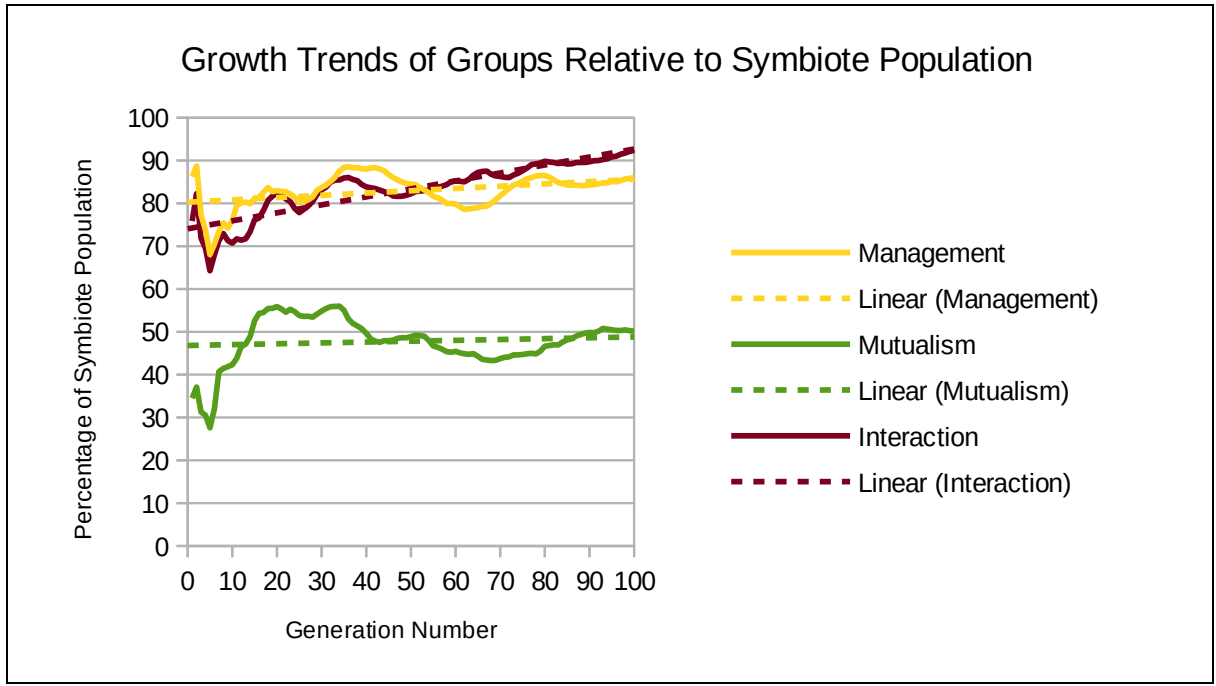

Figure 12. In this graph, we compare the three sets (management, mutualism, and interaction) relative to the subset of the population that consists of symbiotes. Each of the three sets is fitted with linear regression, to determine what trends the curves are following over the generations.

When we normalize management, mutualism, and interaction by the growth of all symbiotes, as in Figure 12, the curves are somewhat noisy, so it is not completely clear what is happening. In Figure 12, we have fitted each of the curves with linear regression (the dashed lines), to get a better view of the situation. All three linear regression lines are increasing (positive slope), which suggests that they are selected by evolution. Table 8 gives the 95% confidence intervals for the three curves. The slopes for management and interaction are positive with 95% confidence. The slope for mutualism is positive, but with less than 95% confidence. However, it could be argued that the early generations are very noisy, with a big drop in generation 5 and a big rise in generations 20 to 35, and therefore we should focus on the more stable generations, from 50 to 100. In these later generations, all three groups (management, mutualism, and interaction) have positive slopes with 95% confidence.

Table 8. When we focus on the second half of the plot in Figure 12 (generations 50 to 100), there is 95% confidence that all three sets (management, mutualism, and interaction) are increasing faster than the set of symbiotes as a whole. This suggests that there is an evolutionary advantage to management, mutualism, and interaction.

| Groups | Linear regression slopes relative to symbiote population | | | | | |
| | Generations 0 to 100 | | | Generations 50 to 100 | | |
| | Lower 95% | Middle | Upper 95% | Lower 95% | Middle | Upper 95% |
|---|---|---|---|---|---|---|
| Management | 0.028 | 0.053 | 0.078 | 0.069 | 0.106 | 0.143 |
| Mutualism | -0.019 | 0.020 | 0.059 | 0.052 | 0.092 | 0.132 |
| Interaction | 0.167 | 0.186 | 0.205 | 0.168 | 0.181 | 0.194 |

We conclude that there is good evidence that evolution selects for management, mutualism, and interaction in Model-S. It appears that Model-S is consistent with the discoveries that biologists have made in studying symbiosis in nature (Douglas, 2010).

# 5 Future Work and Limitations

In this article, we have used Model-S to test three hypotheses that were suggested by the research of biologists who study symbiosis (Douglas, 2010). A closer study of this body of biological research is likely to turn up other hypotheses that could be tested using Model-S. Biologists may also discover ways in which Model-S fails to account for the observations of biologists. These observations could lead to better computational models of symbiosis, and better models could also lead biologists to new discoveries.

We have used three different cellular automata in our work (see Section 2). Although the Management Game was specifically designed for studying symbiosis, it seems unlikely that it is the best cellular automaton for studying symbiosis, given the huge space of possible cellular automata (Turney, 2021b). Even if we restrict our attention to Conway's Game of Life and the various ways colours can be added to the game, we are faced with an infinite number of possible variations.

In Section 3, we briefly mentioned fitness based on competition versus fitness based on growth in isolation. Our focus in this paper has been fitness based on competition, but an interesting question for future work is whether management, mutualism, and interaction are predictive of fitness in isolation, as measured by the rate of growth or the total growth of an isolated organism.

This project was inspired by research on the major cooperative transitions in evolution. The major cooperative transitions in biological and cultural evolution may be seen as cases of particularly successful symbioses. Attempts to explain the processes that drive the major cooperative transitions include the work of Buss (1987), Maynard Smith (1988), Maynard Smith and Szathmáry (1995), Stewart (1995; 1997; 2014;

2020), Michod (1999), Wilson and Wilson (2007), Wilson (2015), West et al. (2015), Szathmáry (2015), and Wilson (2019). In future work, we plan to investigate how Model-S might be adapted as a model of human societies, where humans in a social group are treated as analogous to partners in a symbiote.

# 6 Conclusion

Model-S evolves a population of seed patterns using a fitness measure based on competitive growth. Seed patterns compete in the Immigration Game, where the winner is the pattern that grows the most. The fitness of a seed pattern in Model-S is measured by the average number of games that it wins, competing against all of the other seeds in the population. The experiments in Section 4 show that evolutionary selection in Model-S favours symbiotes with one manager, zero outsiders, and zero soloists. These results are consistent with what we would expect, based on the research of biologists who study symbiosis.

More generally, Model-S shows that cellular automata combined with an evolutionary algorithm can yield computational models of symbiosis that are consistent with the discoveries of biologists. We expect that computational models of this kind will be useful for many aspects of biological research, and perhaps for many aspects of research in models of human cultural development.

# Acknowledgments

Thanks to Andrew Trevorrow, Tom Rokicki, Tim Hutton, Dave Greene, Jason Summers, Maks Verver, Robert Munafo, Brenton Bostick, and Chris Rowett, for developing the Golly cellular automata software. Thanks to John Stewart for encouragement and inspiration. Thanks to Martin Brooks for helpful comments. Thanks to the reviewers for their very helpful comments and suggestions, which have significantly improved this article.